\documentclass{bmvc2k}

\usepackage{algorithm}
\usepackage{algorithmic}
\usepackage{array}
\usepackage{caption} 
\DeclareMathOperator*{\argmax}{arg\,max}

\title{Contour-guided Image Completion with\\
 Perceptual Grouping}

\addauthor{Morteza Rezanejad*}{morteza.rezanejad@utoronto.ca}{123}
\addauthor{Sidharth Gupta*}{sid.gupta@mail.utoronto.ca}{2}
\addauthor{Chandra Gummaluru}{chandra.gummaluru@mail.utoronto.ca}{4}
\addauthor{Ryan Marten}{ryan.marten@gmai.com}{2}
\addauthor{John Wilder}{jdwilder@cs.toronto.edu}{1}
\addauthor{Michael Gruninger}{gruninger@mie.utoronto.ca}{3}
\addauthor{Dirk B. Walther}{bernhardt-walther@psych.utoronto.ca}{1}

\addinstitution{
Department of Psychology\\
University of Toronto\\
Toronto, Canada
}
\addinstitution{
Department of Computer\\ 
Science, University of Toronto\\
Toronto, Canada
}
\addinstitution{
Department of Mechanical \& \\
Industrial Engineering\\
University of Toronto\\
Toronto, Canada
}
\addinstitution{
Department of Electrical and\\
Computer Engineering\\
University of Toronto\\
Toronto, Canada
}

\runninghead{}{}

\begin{document}
%

\maketitle
\vspace{-0.6cm}
\begin{abstract}
Humans are excellent at perceiving illusory outlines. We are readily able to complete contours, shapes, scenes, and even unseen objects when provided with images that contain broken fragments of a connected appearance. In vision science, this ability is largely explained by perceptual grouping: a foundational set of processes in human vision that describes how separated elements can be grouped. In this paper, we revisit an algorithm called Stochastic Completion Fields (SCFs) that mechanizes a set of such processes -- good continuity, closure, and proximity -- through contour completion. This paper implements a modernized model of the SCF algorithm, and uses it in an image editing framework where we propose novel methods to complete fragmented contours. We show how the SCF algorithm plausibly mimics results in human perception. We use the SCF completed contours as guides for inpainting, and show that our guides improve the performance of state-of-the-art models. Additionally, we show that the SCF aids in finding edges in high-noise environments. Overall, our described algorithms resemble an important mechanism in the human visual system, and offer a novel framework that modern computer vision models can benefit from.
\end{abstract}
\vspace{-0.6cm}
\section{Introduction}
\label{sec:intro}
Perceptual grouping is the process of grouping small visual elements, such as edges, into larger, more meaningful structures. Perceptual grouping has a long history in computer vision \cite{witkin1983role,lowe1985perceptualOrganization,sarkar1993perceptual}, but with the advent of deep learning, it has been overlooked in the context of modern computer vision tasks in the past decade. This is possibly because such rules are not easily implemented into modules that can be readily integrated within end-to-end deep learning systems. 
In this paper, we propose a modernized framework that mechanizes classical ideas in perceptual grouping, and address the aforementioned limitation by offering novel ways to integrate perceptual grouping principles in modern computer vision systems. In perceptual grouping, Gestalt psychologists have proposed qualitative grouping principles that govern how the human visual system groups visual content \cite{koffka1922perception}. These principles include proximity, good continuation, symmetry, and closure. While these ideas started as qualitative rules, several concrete and quantitative algorithms are continually being implemented as models of human visual system functions \cite{elder1993effect,wagemans2012century,Wagemans12II,jakel2016overview,wilder2019local,rezanejad2019gestalt}. The same principles have formed the basis for many computer vision algorithms as well \cite{elder1996computing,siddiqi1999shock,rezanejad2015view,levinshtein2013multiscale,lee2015learning,rezanejad2015robust}.
\begin{figure*}[t]
	\begin{center}
		\includegraphics[width=\textwidth,trim=0in 0in 0in 0in,clip]{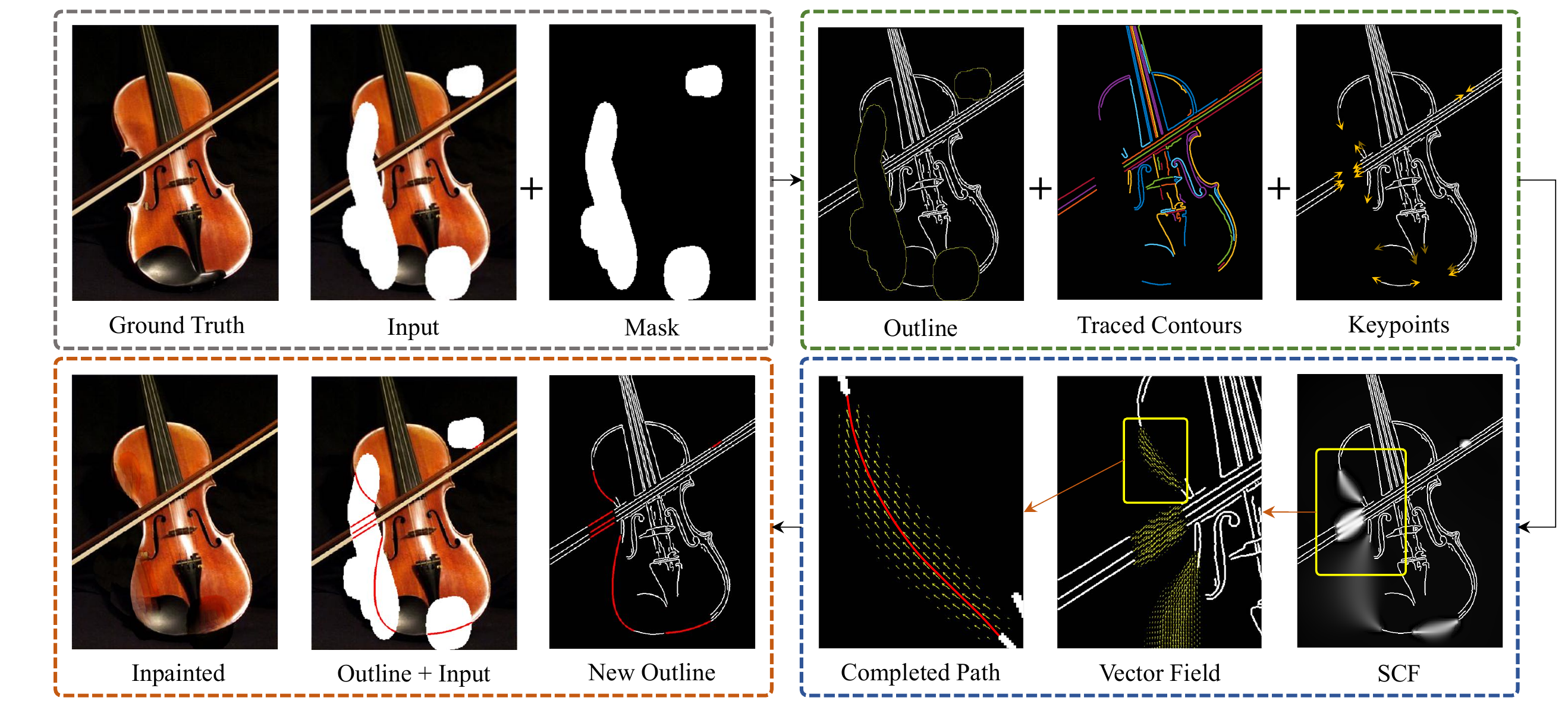}
  \end{center}
	\caption{Here is an example overall view of our stochastic completion field (SCF) processing pipeline used in an image inpainting model. We begin with an original image. First we detect the edges and trace them into individual contours (shown in different colors). Next, we find the keypoints (points to be connected; denoted sources and sinks) and the orientations at each keypoint. These orientations will be the starting directions of the random walks in the SCF. After completing the walks, we can determine the probability of a path going through each point; the brightness in the SCF image represents such probability. Zooming in, we show a vector field that represents the most probable orientation at each point, with the vector magnitude representing probability. We then find the most probably path through the $(x,y,\theta)$ space from one keypoint (source) to the next (sink). We show the combination of the completed path with the original outline and a new, complete outline. We also show the completion over the original masked image, and improved result from inpainting.}
	\label{pipeline}
\end{figure*}
The effortlessness nature of implicit knowledge extraction by deep neural networks paved the way for scientists to focus less on computationally complex models designed for perceptual grouping, simply due to the fact that these models lack the ability to be integrated in modern deep-net systems. However, such frameworks can still benefit and learn cues that are perceptually motivated. \cite{verbin2020field,camaro2020appearance,rezanejad2020medial} have shown that their perceptual grouping framework ``Field of junctions" detects edges better than any state-of-the-art edge detection CNNs, especially in the presence of a large amount of noise. Perceptual grouping-based salience measures can also be combined with neural networks to aid in scene categorization \cite{rezanejad2019scene}. Both of these methods achieve boosts in performance without any additional training, and are integrated within larger computer vision systems. Here, we propose making use of a perceptual grouping framework, \emph{the stochastic completion field} (SCF), to help group contour elements in an image by filling in the gaps between the contours \cite{williams1997stochastic,williams1997local,williams1999computing,momayyez20093d}. We build upon existing frameworks \cite{williams1996local,williams1997local} that compute probability density maps for incomplete contours that can potentially be connected. We revise and modernize the algorithm to achieve contour completion on real appearance images and then use that to address some of the most challenging problems in computer vision. Our contributions are three-fold: 1) We revive and improve the original SCF algorithm in two ways: first by removing the need to assign labels; and second by defining a data structure that allows us to complete contours. 2) We show that our SCF algorithm quantitatively models perceptual grouping processes, and produces results consistent with human behavioural studies. 3) We show that our SCF framework can be integrated in any computer vision model that uses contour shape structures -- be it deep learning or classical (see Fig. \ref{pipeline} for an example of SCF used in inpainting). Specifically, we integrate our SCF algorithm with models for inpainting and edge-detection in noisy environments, and find that our methods significantly improve performance in those tasks. 
Our software framework is made open-source and is available on GitHub (\url{https://github.com/sidguptacode/Stochastic_Completion_Fields}).
\vspace{-0.2cm}
\section{Our Approach to Contour Completion} 
\label{sec:2}
\begin{figure}[t]
     \centering
        \includegraphics[width=\textwidth]{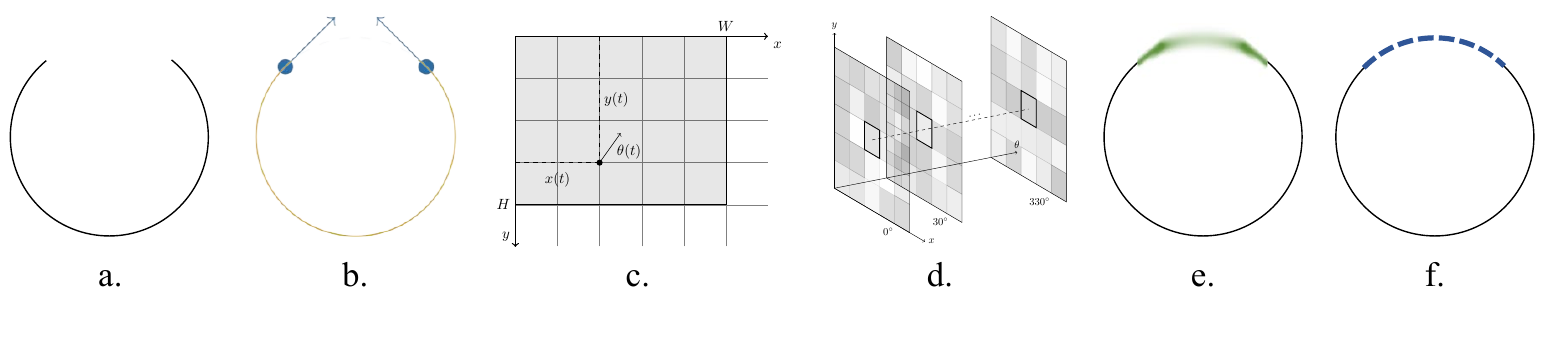}  
     \caption{(a): Fragment figure.  (b): Keypoints of fragment with tangents. (c):Particle on a random walk in $[0, W] \times [0, H]$ at time $t$. (d): For each position, $\theta^*$ is the orientation which results in the maximum field value across all orientations. (e) SCF applied to the source and sink points. (d) Contour traced as the maximum probability path along the SCF.}
     \label{fragmentsAndOrientations}
\end{figure}

In this section, we will address the problem of completing fragmented contours. Let's assume an incomplete contour fragment (e.g., the one shown in Figure~\ref{fragmentsAndOrientations}a) that we want to complete. 
We will first find the keypoints on this fragmented contour, which are the points of interest that need to be connected. Keypoints can be identified in many ways; as the intersection of masks and the original photograph; as where contours of an object or a scene meet the image boundary; where an object is occluded by another object; or in general any start and end point of a contour that is perceptually perceived by the visual input. We can find these keypoints and their tangential orientations (see Fig. \ref{fragmentsAndOrientations}b) using an independent model (such as an edge map generator + logical linear operator \cite{iverson1995logical}). From here, we seek to compute a probability density function called the stochastic completion field (SCF) as shown in Figure \ref{fragmentsAndOrientations}c-e. This SCF will then pave the way to extract a completion path as shown in Figure \ref{fragmentsAndOrientations}f.
\vspace{-0.2cm}
\subsection{The Stochastic Completion Field}
\begin{figure}[t]
    \centering
    \includegraphics[width=0.95\textwidth]{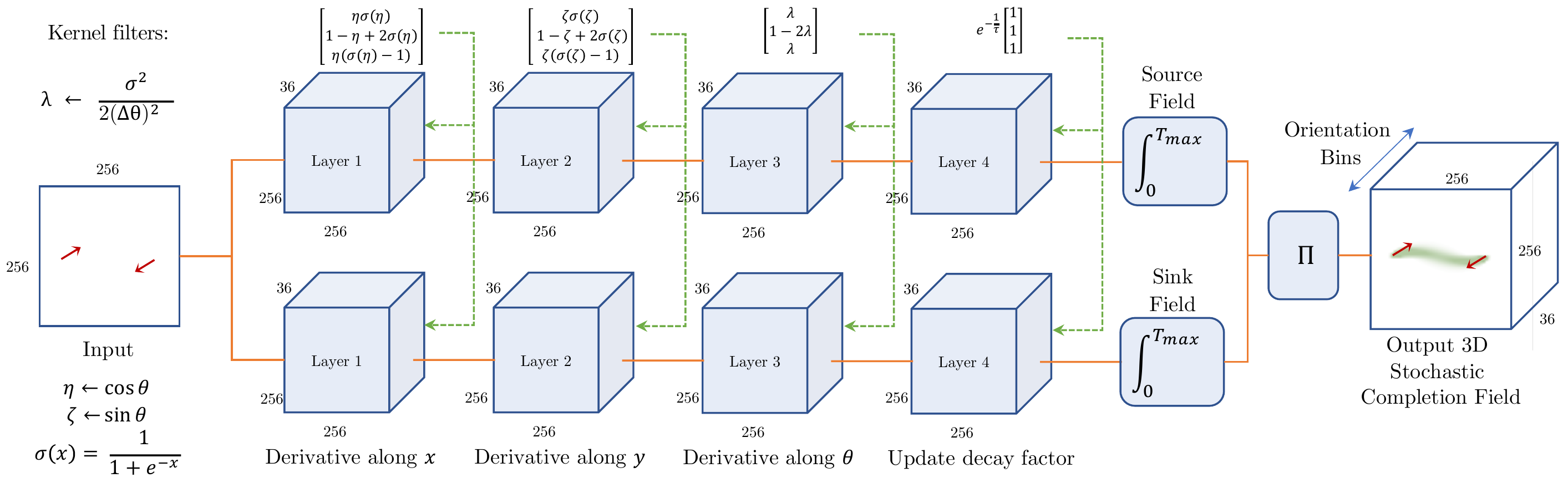}  
    \vspace{5mm}
     \caption{A schematic representation of how the SCF can be computed using a neural network model. Given a source and sink, the neural network computes two 3D tensors: ``source field'' and ``sink field''. Each layer applies one step of the Fokker-Plank algorithm (see Alg. \ref{alg_fokker_plank}). The recurrence relationship given by the Fokker-Plank equations can be computed through the kernel filter convolutions given in this figure.} 
     \label{fig:neural_net}
\end{figure}

The likelihood of a candidate contour being the completion between a pair of keypoints is related to the likelihood of a particle starting at one of those keypoints, called the source, and traveling along the contour before stopping at the other keypoint, called the sink. To compute "one" completion path in this space, we consider a particle that travels from any source to any sink on a random walk in a 2-dimensional region that contains all of the keypoints. Without loss of generality, we take this region to be $[0, W] \times [0, H]$ (see Fig.~\ref{fragmentsAndOrientations}c). The state of the particle at time $t$ is defined by a 3-tuple $(x(t), y(t), \theta(t))$, where ($x(t)$ and $y(t)$) defines the particle's position, and $\theta(t)$ defines it's direction of linear velocity. Assuming that the particle's linear velocity is a unit vector, we can model the change in it's position as $\dot{x}(t) = \cos{\left(\theta(t)\right)} \text{ and } \dot{y}(t) = \sin{\left(\theta(t)\right)}$, with a change in orientation $\dot{\theta}(t)$ that is sampled from a normally distributed stochastic process $\mathcal{N}(0, \sigma^2)$ (with some constant $\sigma$). This procedure allows for a prior expectation of a smooth, unbroken set of completion contours, or in other words, contours with ``good continuation''. Now, we can represent the \textit{stochastic completion field} by a function $C: [0, W] \times [0, H] \times [0, 2\pi) \rightarrow[0,1]$, such that $C(s)$ gives the likelihood of being in a state $s$, when moving from any source to any sink. If we let  $p(u,s,v)$ be the likelihood that a random walk starts at $u$, passes through $s$ and ends at $v$, then C(s) can be written as $ \iint p(u,s,v)dudv$. Next, we expand $p(u,s,v)$ by introducing four components:
1) the likelihood that $u$ is the source of a random walk, denoted $p_{\mathcal{U}}(u)$; 2) the likelihood that a random walk begins at $u$ , ends at $s$, denoted $p(s|u)$; 3) the likelihood that a random walk begins at $u$ and passes through $s$, ends at $v$, denoted $p(v|u,s)$; 4) the likelihood that $v$ is a sink, denoted, $p_{\mathcal{V}}(v)$. A random walk is a Markov process, meaning a future state is only influenced by the current state and nothing before. This means that the likelihood $p(v|u,s)$ is independent of the starting point $u$ and only depends on the current state $s$, so $p(v|u,s) = p(v|s)$. Therefore, we can compute the $p(u,s,v)$ by separating the conditionals on $u$ and $v$. Specifically, this means that the SCF can be computed as  $C(s) = U(s)V(s)$ where $U(s)$ and $V(s)$ are probability maps defined as: $U(s) = \int{p_{\mathcal{U}}(u)p(s|u)du}$ for the \textit{source field} and $V(s) = \int{p_{\mathcal{V}}(v)p(v|s)dv}$ for the \textit{sink field}. We can evaluate $C(s)$ by first computing the conditional distribution at a given time, $t$, which we denote as $C(s;t)$, and then marginalizing over all $t$. Using the same principles, $C(s;t) = U(s;t)V(s;t)$ where $U(s;t) = \int{p_{\mathcal{U}}(u)p(s|u;t)du}$ and $V(s;t) = \int{p_{\mathcal{V}}(v)p(v|s;t)dv}$. Both the source field ($U(s;t)$) and the sink field ($V(s;t)$) can be computed in the same way, because if $u$ and $v$ are swapped, we will achieve the same distribution. Now, we will simplify our computations by detailing how these fields can be computationally modeled. Let us imagine that we are computing a source / sink field, and $P(x,y,\theta,t)$ represents the likelihood of a state $s = (x, y, \theta)$ at time $t$. This density $P$ can be computed from the initial condition ($t=0$) with the following Fokker-Planck equation \cite{williams1996local}
$P(x,y,\theta,t) = P(x,y,\theta,0) + \int_{0}^{t}\frac{\partial P(x,y,\theta,t')}{\partial t}dt',$ and$\frac{\partial P}{\partial t} = -\cos{\theta}\frac{\partial P}{\partial x} - \sin{\theta}\frac{\partial P}{\partial y} + \frac{\sigma^2}{2}\frac{\partial^2P}{\partial\theta^2} - \frac{1}{\tau}P$
where  $P(x, y, \theta;t)$ represents the probability that a particle is at position $(x, y)$ with orientation $\theta$ at time $t$. The term $\tau$ is a constant particle decay rate which enables a prior expectation of ``proximity'', i.e., ensuring that the likelihood of being on a path gets exponentially smaller as we get further away from a source or sink.
\begin{algorithm}[tb]
\caption{Fokker-Plank($x, y, \theta, \tilde{P}_{\text{init}}, T_{\text{max}}$)}
\label{alg_fokker_plank}
\begin{algorithmic}
\STATE $\lambda \leftarrow \sigma^2 / 2(\Delta\theta)^2$
\FOR{$t = 0, \dots, T_{\text{max}}$}
\STATE $\tilde{P}^{t}_{x,y,\theta} \leftarrow \tilde{P}^{t}_{x,y,\theta} - \cos{\theta}\cdot \tilde{P}^t_{x,y,\theta}+\cos{\theta}\cdot\tilde{P}^t_{x-\Delta x,y,\theta} \text{ if } \cos{\theta} \geq 0$ \COMMENT{Derivative along $x$}
\STATE  $\tilde{P}^{t}_{x,y,\theta} \leftarrow \tilde{P}^{t}_{x,y,\theta} - \cos{\theta}\cdot \tilde{P}^t_{x+\Delta x,y,\theta}+\cos{\theta}\cdot\tilde{P}^t_{x,y,\theta} \text{ if } \cos{\theta} < 0$ \COMMENT{Derivative along $x$}
\STATE $\tilde{P}^{t}_{x,y,\theta} \leftarrow\tilde{P}^{t}_{x,y,\theta} - \sin{\theta}\cdot
\tilde{P}^{t}_{x,y,\theta}+ \sin{\theta}\cdot\tilde{P}^{t}_{x,y-\Delta y,\theta}
\text{ if } 
\sin{\theta} \geq 0$ \COMMENT{Derivative along $y$}
\STATE  $\tilde{P}^{t}_{x,y,\theta} \leftarrow \tilde{P}^{t}_{x,y,\theta} - \sin{\theta}\cdot
\tilde{P}^{t}_{x,y+\Delta y,\theta}+\sin{\theta}\cdot\tilde{P}^{t}_{x,y,\theta}
\text{ if } 
\sin{\theta} < 0$ \COMMENT{Derivative along $y$}
\STATE $ \tilde{P}^{t}_{x,y,\theta} \leftarrow \lambda \tilde{P}^{t}_{x,y,\theta-\Delta \theta}+(1-2\lambda)\tilde{P}^{t}_{x,y,\theta}+\lambda \tilde{P}^{t}_{x,y,\theta+\lambda \theta} $ \COMMENT{Derivative along $\theta$}
\STATE $\tilde{P}^{t+1}_{x,y,\theta} \leftarrow \exp \left\lbrace{-\frac{1}{\tau}}\right\rbrace \cdot \tilde{P}^{t}_{x,y,\theta}$ \COMMENT{Update decay factor}
\ENDFOR
\end{algorithmic}
\end{algorithm}
The solution to the Fokker-Planck equation can be approximated using a finite difference method derived by taking the first-order terms of a Taylor series. We consider $W/\Delta x$ evenly spaced points over $[0, W]$, $H/\Delta y$ evenly spaced points over $[0, H]$, and $2\pi / \Delta \theta$ evenly spaced angles over $[0, 2\pi)$. We then define a function, \textsc{Fokker\_Plank}$(x,y,\theta, T_{\text{max}})$, that approximates $P(x,y,\theta,t)$ for a discrete state, $(x,y,\theta)$ and all $t \leq T_{\text{max}}$. Its implementation is described in Alg. \ref{alg_fokker_plank}.  
Applying \textsc{Fokker\_Plank} to each state, we obtain $\tilde{P}$, the finite difference approximation of $P$ for $t \leq T_{\text{max}}$. We can then readily approximate $p(s|u;t)$ and $p(v|s;t)$ by choosing $\tilde{P}_{\text{init}}$ to be the discrete approximations of $p_{\mathcal{U}}$ and $p_{\mathcal{V}}$, respectively. From here, computing $C$ becomes a matter of integration. 
Finally, we can do all these computations using a neural network model with weights that are updated by the sampling function for orientation and the decay factor, as shown in Fig. \ref{fig:neural_net}.
\subsection{Assigning Sources and Sinks}
Generating an SCF requires labelling keypoints as either sources or sinks, since we need to know the specific pairs of keypoints that should be connected. The original formulation of SCFs assumes that this labelling is given, but that is most often not the case when we see broken contour fragments. We propose an algorithm that labels keypoints as sources or sinks automatically, letting us know which points should be connected. Our approach is to marginalize the distribution over all possible assignments of sources and sinks. We define a loop, where during the $i$\textsuperscript{th} iteration, we compute an SCF by assigning the $i$\textsuperscript{th} keypoint as the source, and the remaining keypoints as sinks. Intuitively, this generates a probability distribution modeling the path from that specific source and visiting any of the remaining keypoints. We accumulate the SCFs computed at each iteration into one SCF.
\subsection{Tracing Optimal Completion Paths}
\label{sec:optimal_completion_path}
We now describe our approach for tracing the optimal completion paths given an SCF. First, for each discrete position, $\mathbf{p} = (x,y)$, we find the discrete orientation, $\theta^*$, which maximizes the field value, that is, $\theta^* = \argmax_{\theta}{\tilde{C}(x,y,\theta)}$. $\mathbf{p}$ is then assigned a vector whose magnitude is $\tilde{C}(x,y,\theta^*)$ and whose direction is $\theta^*$. The result is a discrete vector field, $\tilde{C}_{\text{vec}}(x,y)$, that gives the most likely instantaneous direction of motion of the particle at $(x,y)$. Given a starting point $(x_1, y_1)$ and end point $(x_2, y_2)$, we can now trace the most optimal path by greedily taking steps of a particular size in these most probable directions. As the resulting positions may no longer be on the discrete grid, we use a linear interpolation of $\tilde{C}_{\text{vec}}$ over $[0, W] \times [0, H]$ to compute the corresponding vector on the new position from the vector field.
The algorithm iteratively builds a list of points representing our traced path as described, and it terminates once the current position is near the specified end point (within a distance radius of a particular value). Now the result is an automated algorithm that computationally completes fragments from disconnected boundaries, without any training. The applications of such a toolbox, as well as its effectiveness, are explored in the following sections.
\begin{figure*}[!tb]
	\begin{center}
		\includegraphics[width=0.95\textwidth,trim=0in 0in 0in 0in,clip]{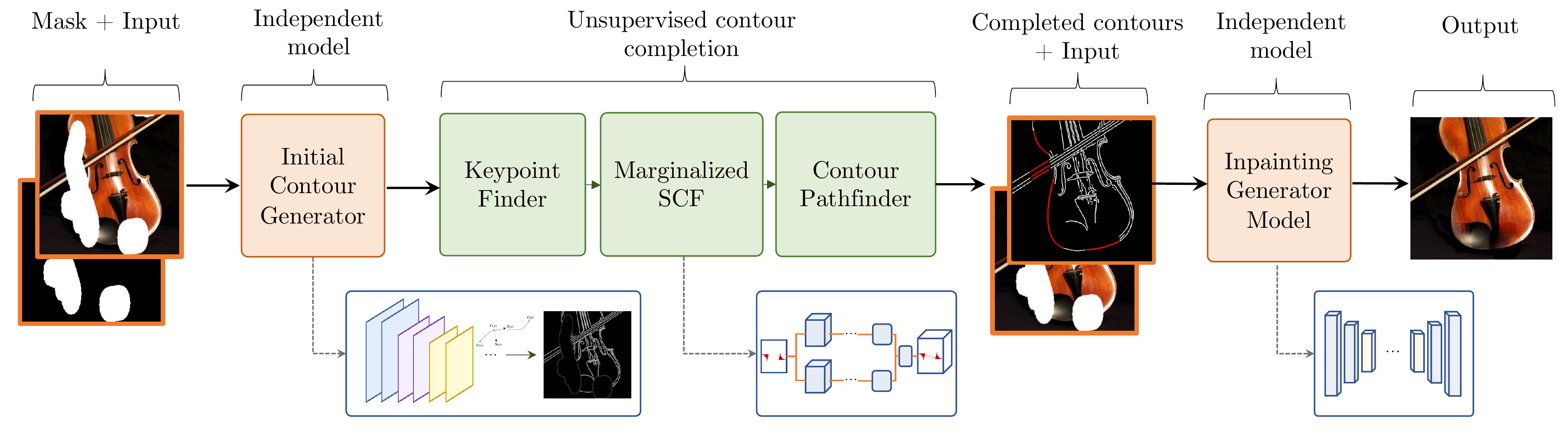}
  \end{center}
	\caption{Pipeline of how SCFs are used in computer vision systems. The orange blocks can represent any computer vision model that uses contour information (in this case, we have a contour generator and an inpainting model). The green blocks show how SCFs integrate into the system for added perceptual grouping information. The Marginalized SCF block can be implemented via Alg. \ref{alg_fokker_plank}, or as the neural network mentioned in Fig. \ref{fig:neural_net}.}
	\label{alg_box}
\end{figure*}
\section{Image Inpainting}

Image inpainting is the process of recovering content from an image with missing regions. These missing regions are represented as pixels without actual image information, and hence there is uncertainty of the content. 
Recently, it has been demonstrated that modern inpainting methods can be improved through guiding the inpainting models with added geometric constraints \cite{Yu_2018_CVPR, xiong2019foreground, Xiong_2019_CVPR}. \cite{yu2019free} found that by adding a "guide" to show where there is a boundary between regions of distinct image content, the inpainted images are rated as more similar to the intact images by human observers. EdgeConnect \cite{nazeri2019edgeconnect} take this further by trying to ``hallucinate'' missing edges, which yields images with more boundaries, and allows the generated inpainted images to retain finer details. Here, we show that completion of missing contour fragments can play a big role in solving problems of this nature. Concretely, we use SCFs to generate contours that will serve as guidance for inpainting models. 
We first complete the contours through a masked region, as described in section \ref{sec:optimal_completion_path}. Then, those contours guide the generative inpainting models to fill in the missing context from the original image, with our added geometric constraints (see Fig. \ref{alg_box} for a visual representation of how our inpainting setup is implemented).
\section{Experiments and Results}
\label{sec:3}
\subsection{Object Boundary Completion}
\begin{figure}[t]
	\begin{center}
  		\includegraphics[width=0.95\textwidth,trim=0in 0in 0in 0in,clip]{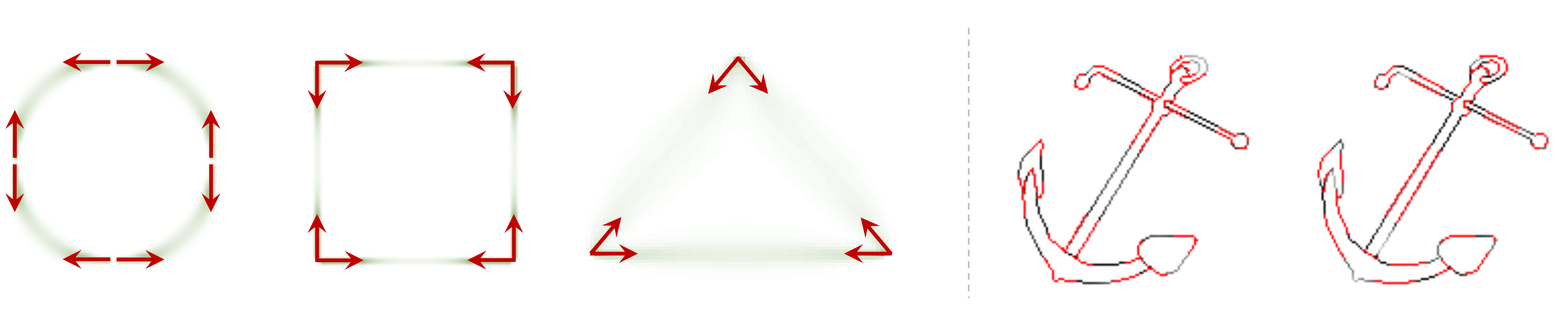}
  \end{center}
	\caption{Examples of stochastic completion fields. Probability is denoted by luminance (dark = high probability). Near the sources and sinks there is high probability, and as we get farther from the keypoints there is less probability. Probability is averaged over all orientations of the completion field. Right: example completion fields for half-image line drawings of an anchor. The original line drawings are from \cite{snodgrass1980standardized}. The left image shows the completion for the junction half image, and the right shows the completion for the middle half image. We show the dot product with the completed SCF, and the image containing only the missing segments, as described in Section \ref{sec:3}.}
	\label{completionExamples}
\end{figure}

We start this section by showing some ``toy'' examples of SCFs in Fig.~\ref{completionExamples} (left). The three images show the result when three or four pairs of keypoints are positioned next to each other in opposite directions, as if they lie on a circle, on a square or on a triangle.
\cite{biederman1991priming} showed that human observers are better at classifying degraded line drawings when shown only contour junctions than when shown only the middle segments. When subjects are given an incomplete object boundary, their visual systems try to complete the missing boundary fragments as much as possible to make a better prediction of what the representative object class could be. When the visual system has difficulty completing the objects, the object is not recognizable. In this section, we show how the SCF's computational power compares with human behavioural studies. In this experiment, which was motivated by the behavioural experiment of \cite{biederman1991evidence}, we tested the algorithm on the Snodgrass and Vanderwart dataset of 260 manually traced line drawings of objects \cite{snodgrass1980standardized}. The traced objects were separated into half-images, one set with the segments containing junctions, and the other set with the segments containing the pieces between junctions (middle segments). We then completed the two types of half-drawings using the SCF algorithm. To test the quality of the completion, we take the dot product with the completed SCF, and the image containing only the missing segments. This gives a single number representing how well the SCF matches the missing segments.
Our results show that the SCF completes the half-images with the junctions more faithfully than the half-images with the middle segments in 241 / 260 (93\%) of the drawings, aligning with the results of \cite{biederman1991priming} (see Fig. \ref{completionExamples} left). The difference in the quality of the completions was highly significant ($t(259) = 20.64, p < 2.2\times 10^{-16}$). Table~\ref{vss} shows a comparison of the results. 
\begin{table}[b]
\begin{center}
\begin{tabular}{c|cc}
    & Mid-segments & Junctions \\ \hline
    Proportion Correct \cite{biederman1991priming} & \bf{0.833} & 0.727 \\
    SCF dot product a.u. ($\sigma$) & \bf{0.298} (0.49) & 0.152 (0.14)\\
\end{tabular}\vspace{3mm}
\caption{Table comparing the human results \cite{biederman1991priming} and the SCF. The quality of the completions matches how human observers recognize incomplete objects. The human proportion of correct recognition is higher for junctions, just as the dot product between the truth and the completion of the SCF for junctions is higher.}
\label{vss}
\end{center}
\end{table}
This suggests that SCFs are useful for predicting the types of incomplete line drawings that are easy to complete by the human visual system. It also demonstrates that SCFs can form a plausible algorithmic implementation of the Gestalt principle of good continuation.
\begin{figure*}[!htb]
\renewcommand{\arraystretch}{0.3}
\begin{tabular}{c@{\hskip 1pt}c@{\hskip 1pt}c@{\hskip 1pt}c@{\hskip 1pt}c@{\hskip 1pt}c}
    \includegraphics[width=0.164\textwidth,trim=0in 0in 0in 0in,clip]{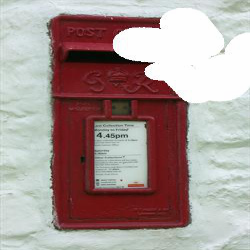} & \includegraphics[width=0.164\textwidth,trim=0in 0in 0in 0in,clip]{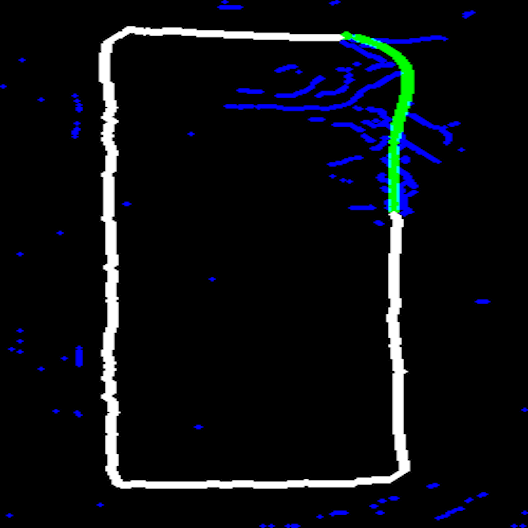} & \includegraphics[width=0.164\textwidth,trim=0in 0in 0in 0in,clip]{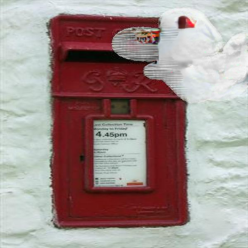} &\includegraphics[width=0.164\textwidth,trim=0in 0in 0in 0in,clip]{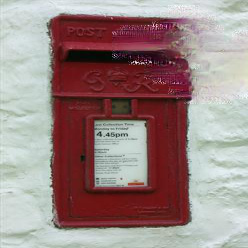} &\includegraphics[width=0.164\textwidth,trim=0in 0in 0in 0in,clip]{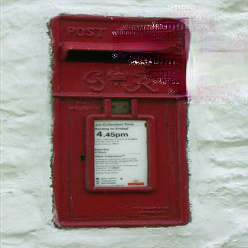} & \includegraphics[width=0.164\textwidth,trim=0in 0in 0in 0in,clip]{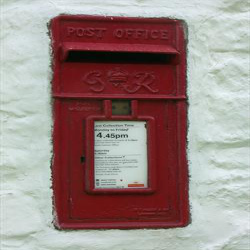}
    \\
    \includegraphics[width=0.164\textwidth,trim=0in 0in 0in 0in,clip]{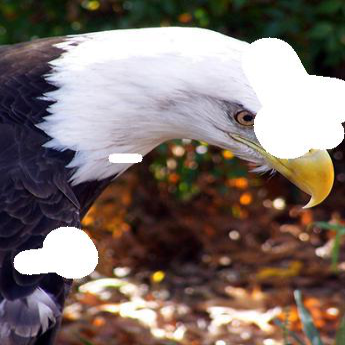} & \includegraphics[width=0.164\textwidth,trim=0in 0in 0in 0in,clip]{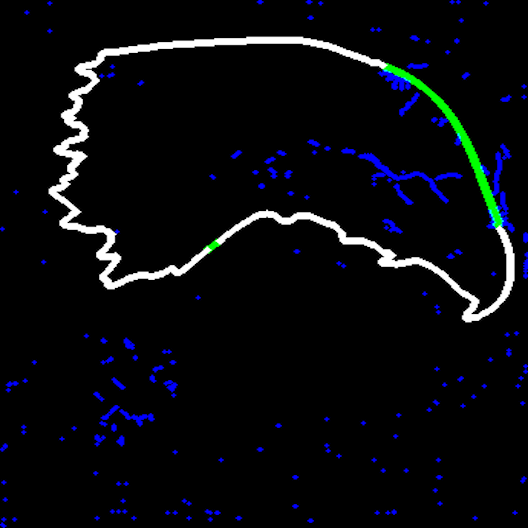} & \includegraphics[width=0.164\textwidth,trim=0in 0in 0in 0in,clip]{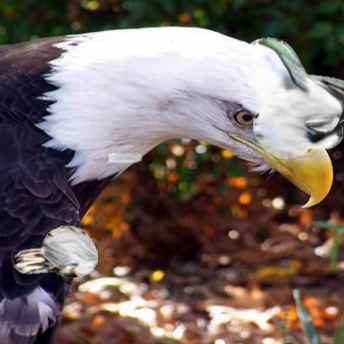} &\includegraphics[width=0.164\textwidth,trim=0in 0in 0in 0in,clip]{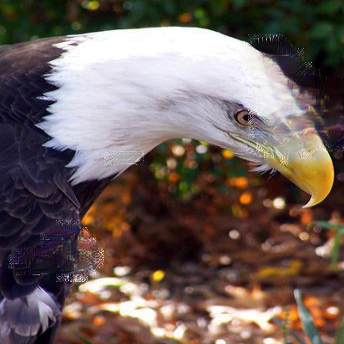} &\includegraphics[width=0.164\textwidth,trim=0in 0in 0in 0in,clip]{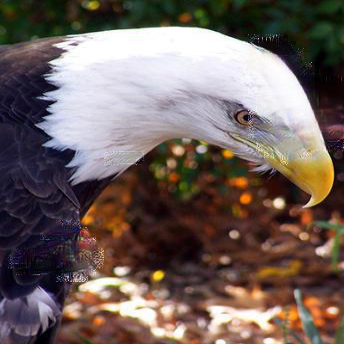} & \includegraphics[width=0.164\textwidth,trim=0in 0in 0in 0in,clip]{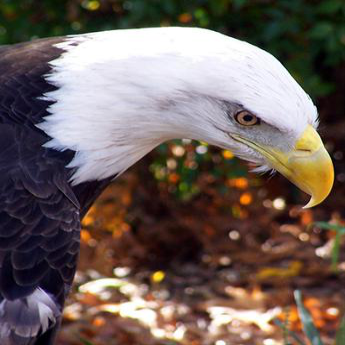}
    \\
     \includegraphics[width=0.164\textwidth,trim=0in 0in 0in 0in,clip]{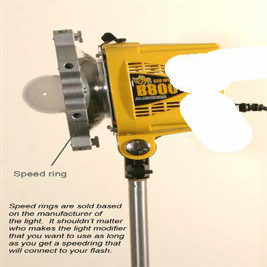} & \includegraphics[width=0.164\textwidth,trim=0in 0in 0in 0in,clip]{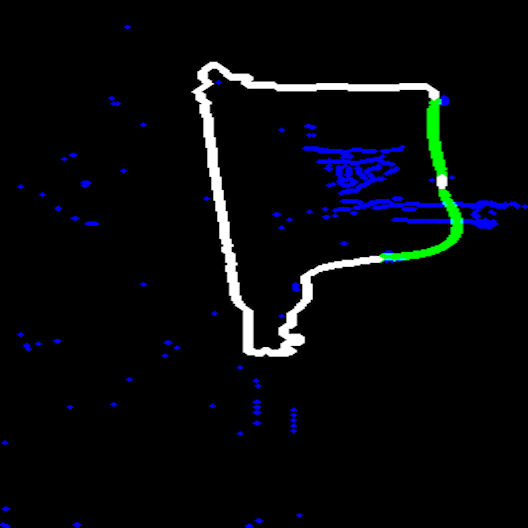} & \includegraphics[width=0.164\textwidth,trim=0in 0in 0in 0in,clip]{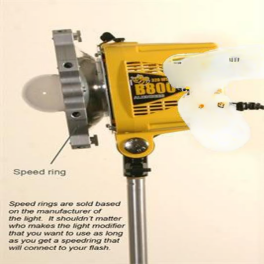} &\includegraphics[width=0.164\textwidth,trim=0in 0in 0in 0in,clip]{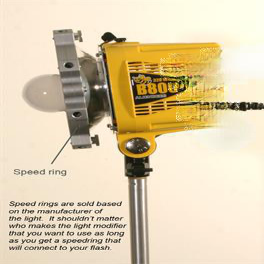} &\includegraphics[width=0.164\textwidth,trim=0in 0in 0in 0in,clip]{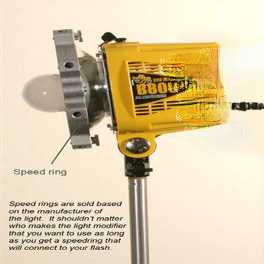} & \includegraphics[width=0.164\textwidth,trim=0in 0in 0in 0in,clip]{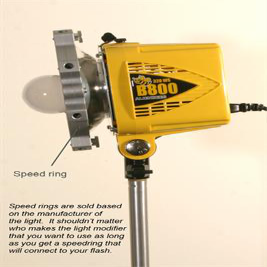}
    \\
    \includegraphics[width=0.164\textwidth,trim=0in 0in 0in 0in,clip]{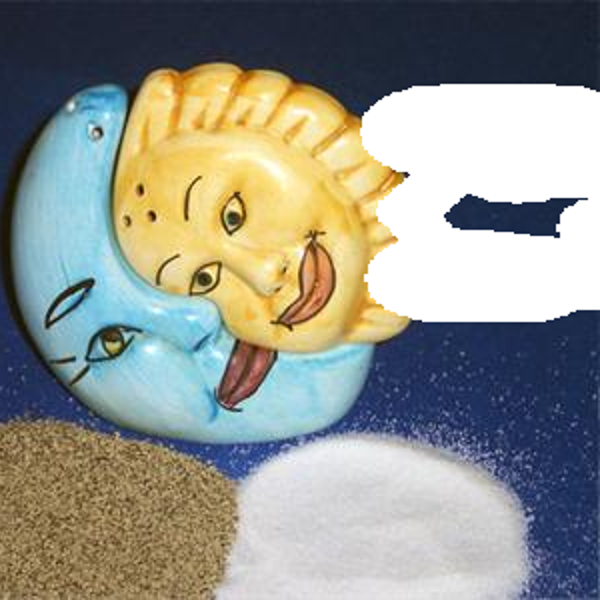} & \includegraphics[width=0.164\textwidth,trim=0in 0in 0in 0in,clip]{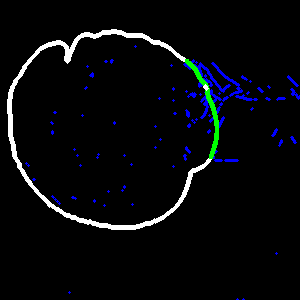} & \includegraphics[width=0.164\textwidth,trim=0in 0in 0in 0in,clip]{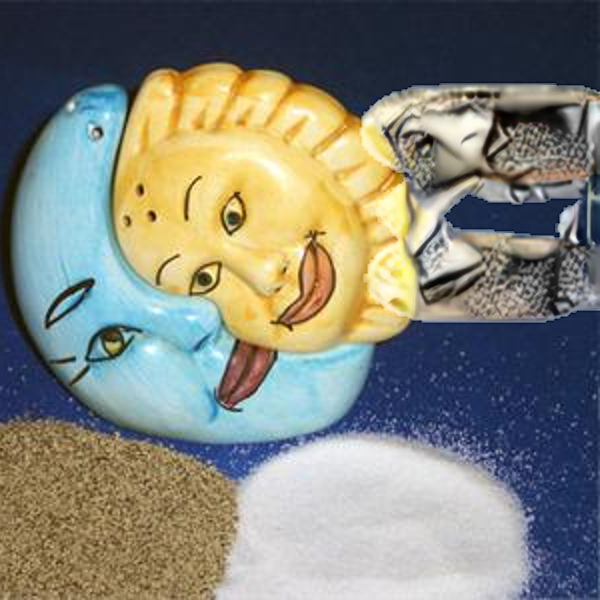} &\includegraphics[width=0.164\textwidth,trim=0in 0in 0in 0in,clip]{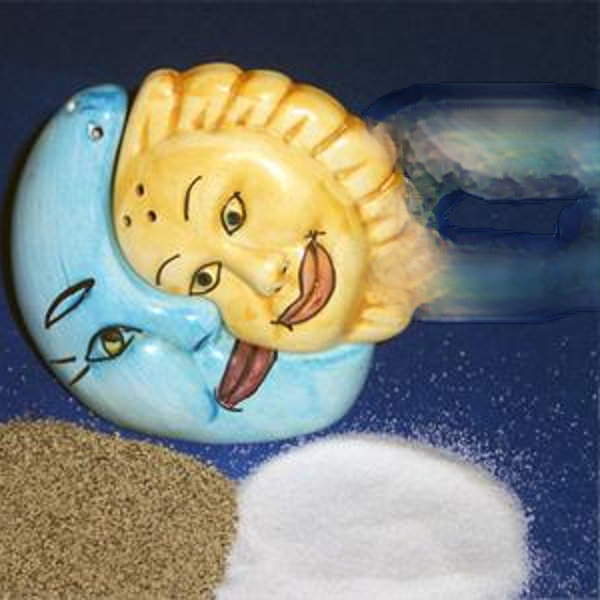} &\includegraphics[width=0.164\textwidth,trim=0in 0in 0in 0in,clip]{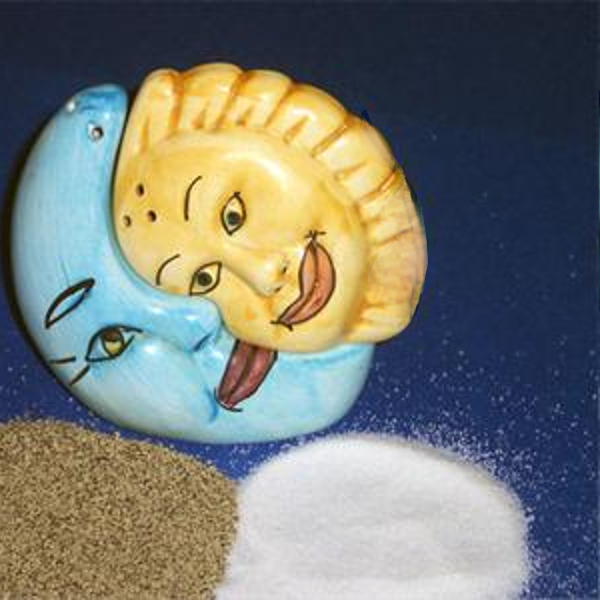} & \includegraphics[width=0.164\textwidth,trim=0in 0in 0in 0in,clip]{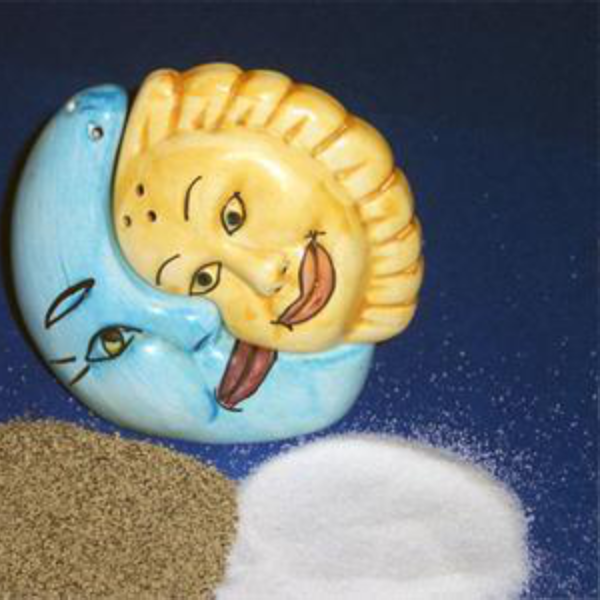}
    \\
    \includegraphics[width=0.164\textwidth,trim=0in 0in 0in 0in,clip]{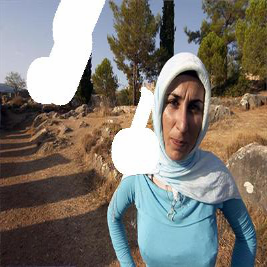} & \includegraphics[width=0.164\textwidth,trim=0in 0in 0in 0in,clip]{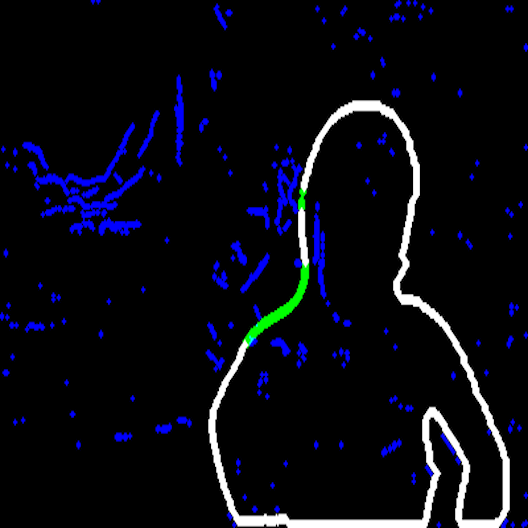} & \includegraphics[width=0.164\textwidth,trim=0in 0in 0in 0in,clip]{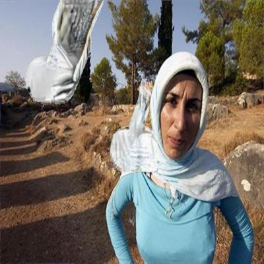} &\includegraphics[width=0.164\textwidth,trim=0in 0in 0in 0in,clip]{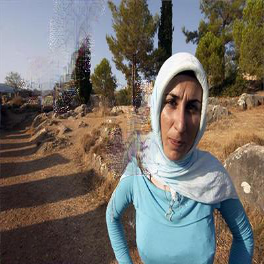} &\includegraphics[width=0.164\textwidth,trim=0in 0in 0in 0in,clip]{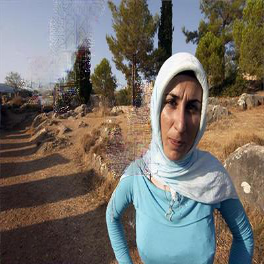} & \includegraphics[width=0.164\textwidth,trim=0in 0in 0in 0in,clip]{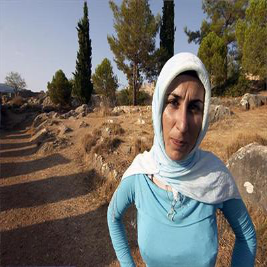}
    \\
    \includegraphics[width=0.164\textwidth,trim=0in 0in 0in 0in,clip]{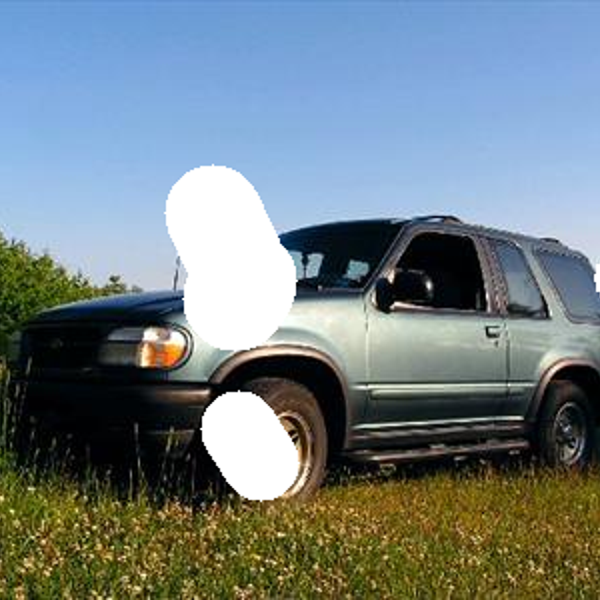} & \includegraphics[width=0.164\textwidth,trim=0in 0in 0in 0in,clip]{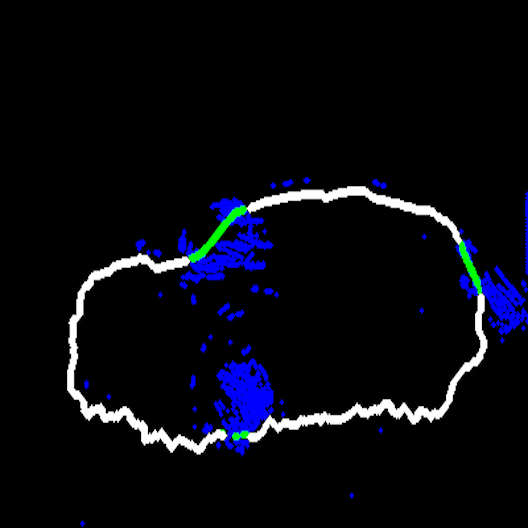} & \includegraphics[width=0.164\textwidth,trim=0in 0in 0in 0in,clip]{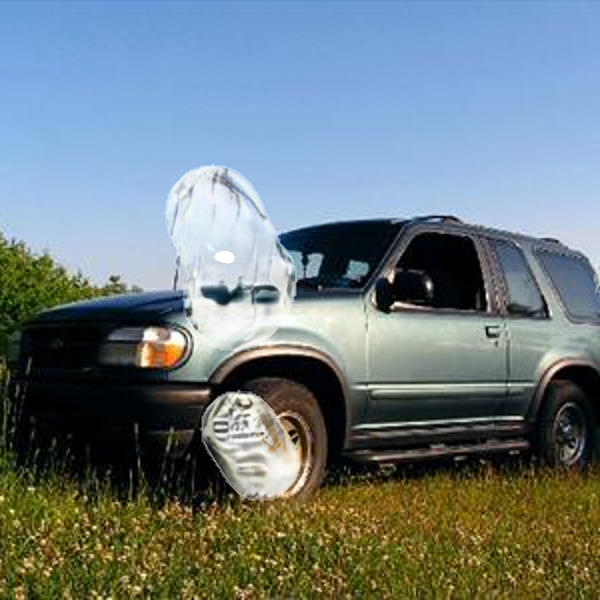} &\includegraphics[width=0.164\textwidth,trim=0in 0in 0in 0in,clip]{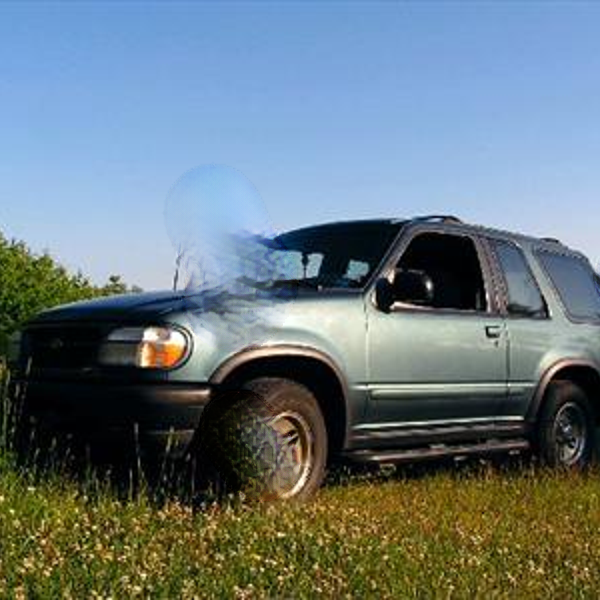} &\includegraphics[width=0.164\textwidth,trim=0in 0in 0in 0in,clip]{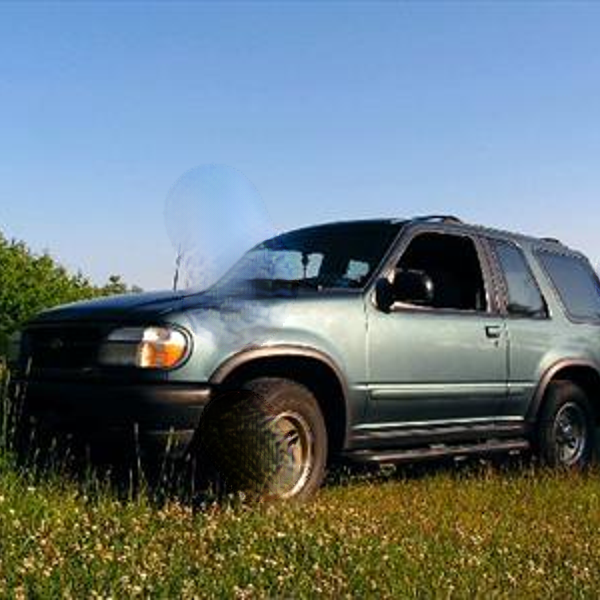} & \includegraphics[width=0.164\textwidth,trim=0in 0in 0in 0in,clip]{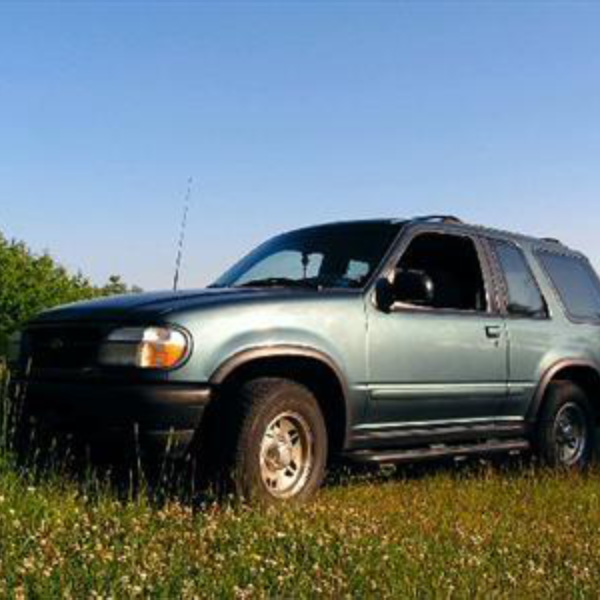}
    \\
     \includegraphics[width=0.164\textwidth,trim=0in 0in 0in 0in,clip]{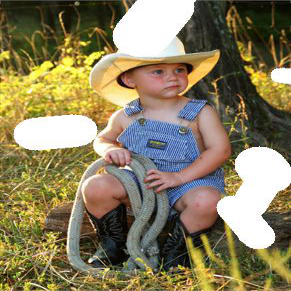} & \includegraphics[width=0.164\textwidth,trim=0in 0in 0in 0in,clip]{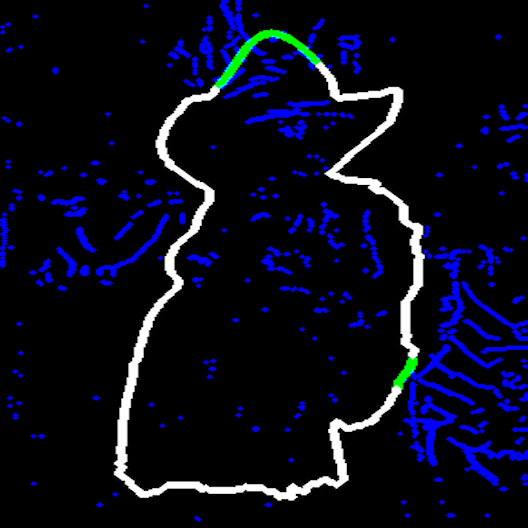} & \includegraphics[width=0.164\textwidth,trim=0in 0in 0in 0in,clip]{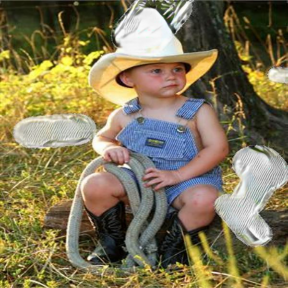} &\includegraphics[width=0.164\textwidth,trim=0in 0in 0in 0in,clip]{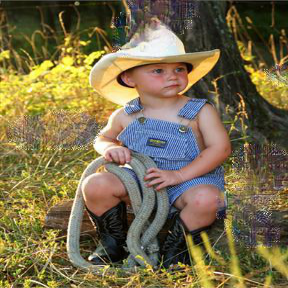} &\includegraphics[width=0.164\textwidth,trim=0in 0in 0in 0in,clip]{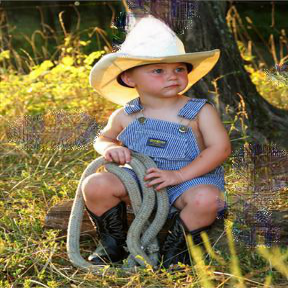} & \includegraphics[width=0.164\textwidth,trim=0in 0in 0in 0in,clip]{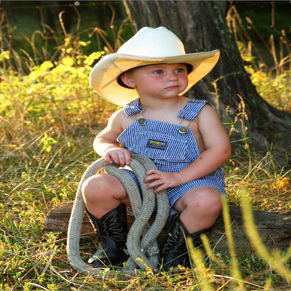}
    \\
    \scriptsize{Input} & \scriptsize{Contours} & \scriptsize{GatedConv} & \scriptsize{EdgeConnect} & \scriptsize{Ours} & \scriptsize{Ground Truth}\\\vspace{1mm}
\end{tabular}
\vspace{0.2cm}
    \centering
    \caption{Results comparing between in-painting methods, including EdgeConnect (EC) that uses contour guidance. In ours, we complete edges using SCF to provide guidance in the place of hallucinated contours generated by EdgeConnect. Please note that in the second column, our contours are shown by ``green'' and contours hallucinated by EdgeConnect are shown in ``blue''.  See supplementary materials for more examples.}
    \label{fig:inpainting}
\end{figure*}
\vspace{3mm}
\subsection{Inpainting Experiments}
We evaluate our image completion pipeline on the MSRA-10K dataset, which is a dataset of objects originally used for object salience detection \cite{HouPami19Dss}.  We show the output of \cite{yu2019free, nazeri2019edgeconnect} and our method, along with the completed contours used for guidance in Figure~\ref{fig:inpainting} for several examples from the MSRA-10k dataset.

In a test set of 1688 different images, we ran edge connect \cite{nazeri2019edgeconnect} using hallucinated edges (with no manual intervention), and also using the same set of edges, this time completed with SCFs. We compare how each of these methods perform by using the structured similarity index measure (SSIM). We found that inpainting using the SCF completions were significantly closer to the ground truth than inpainting using only EdgeConnect's hallucinated edges ($t(1687) = 13.05, p < 3.6\times10^{-37}$). In fact, on 1066 of the 1688 images tested, inpainting with the SCFs as a guide yielded better results than its counterpart alternative (EdgeConnect). 
We compared the quality of the results to other methods, including peak signal-to-noise ratio (PSNR), $\mathit{l}_1$ and $\mathit{l}_2$. PSNR also shows that our results are a significant improvement over EdgeConnect ($t(1687) = 3.06, p = 0.002$), though the effect is less strong under this measure. $\mathit{l}_1$ and $\mathit{l}_2$ show that our method is a highly significant improvement over EdgeConnect ($t(1687) = 29.21, p = 3.8 \times 10^{-152}$, and $t(1687)=14.52, p = 4.3 \times 10^{-45}$ respectively). 
\begin{table}[b]
    \centering
    \begin{tabular}{c|c|c|c}
    Measure  & gPb & DexiNed & Ours \\\hline\hline
    Precision & 0.1258 & 0.4189 & \textbf{0.6895}\\\hline
    Recall & \textbf{0.3130} &  0.1131 & 0.1636\\\hline
    $F1$ & 0.1795 & 0.1781 & \textbf{0.2645}\\
\end{tabular}\vspace{3mm}
    \caption{Results (Precision, Recall and  $F1$) scores for different methods of detecting edges}
    \label{fig:my_label}
\end{table}
\vspace{0.1cm}
\subsection{Detecting Edges in Noise}
\begin{figure*}[!tb]
\begin{center}
\renewcommand{\arraystretch}{0.3}
\begin{tabular}{c@{\hskip 1pt}c@{\hskip 1pt}c@{\hskip 1pt}c@{\hskip 1pt}c@{\hskip 1pt}c@{\hskip 1pt}c}
    \includegraphics[width=0.16\textwidth,height=0.13\textwidth,trim=0in 0in 0in 0in,clip]{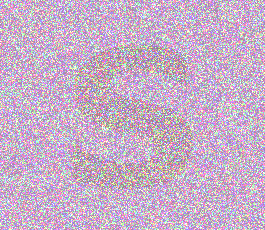} & \includegraphics[width=0.16\textwidth,height=0.13\textwidth,trim=0in 0in 0in 0in,clip]{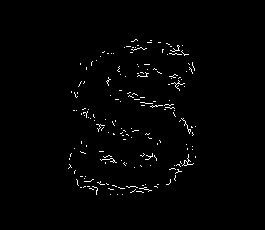} & 
    \includegraphics[width=0.16\textwidth,height=0.13\textwidth,trim=0in 0in 0in 0in,clip]{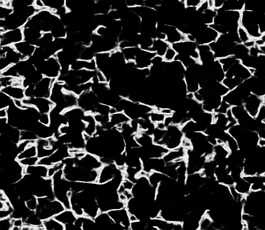} & \includegraphics[width=0.16\textwidth,height=0.13\textwidth,trim=0in 0in 0in 0in,clip]{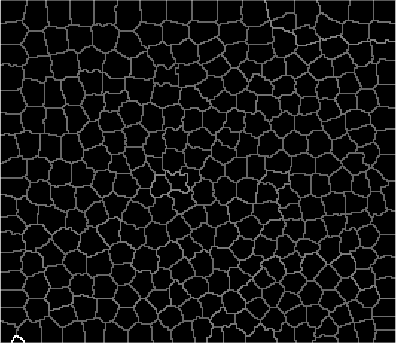} & \includegraphics[width=0.16\textwidth,height=0.13\textwidth,trim=0in 0in 0in 0in,clip]{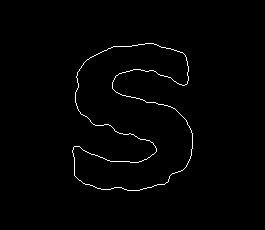} &\includegraphics[width=0.16\textwidth,height=0.13\textwidth,trim=0in 0in 0in 0in,clip]{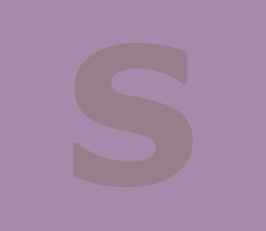} \\
    \includegraphics[width=0.16\textwidth,height=0.13\textwidth,trim=0in 0in 0in 0in,clip]{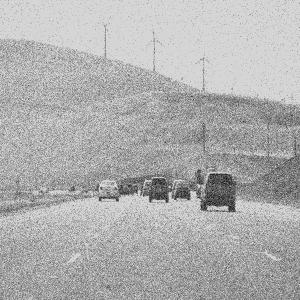} & \includegraphics[width=0.16\textwidth,height=0.13\textwidth,trim=0in 0in 0in 0in,clip]{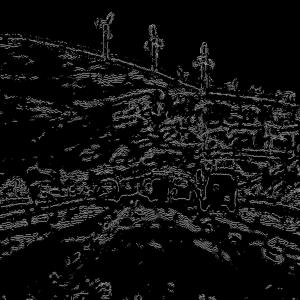} & 
    \includegraphics[width=0.16\textwidth,height=0.13\textwidth,trim=0in 0in 0in 0in,clip]{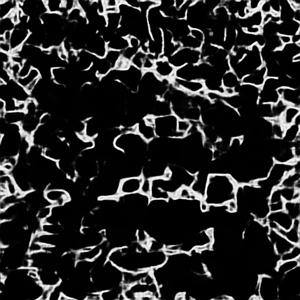} & \includegraphics[width=0.16\textwidth,height=0.13\textwidth,trim=0in 0in 0in 0in,clip]{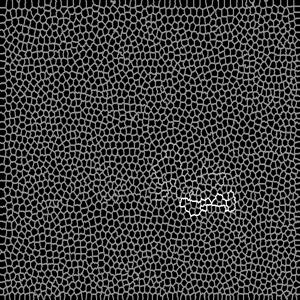} & \includegraphics[width=0.16\textwidth,height=0.13\textwidth,trim=0in 0in 0in 0in,clip]{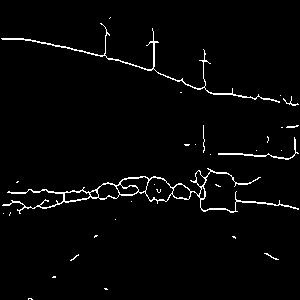} &\includegraphics[width=0.16\textwidth,height=0.13\textwidth,trim=0in 0in 0in 0in,clip]{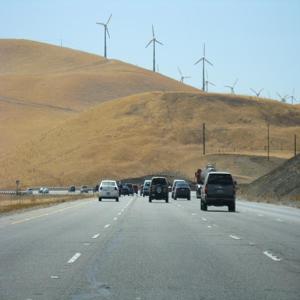}\\
    \includegraphics[width=0.16\textwidth,trim=0in 0in 0in 0in,clip]{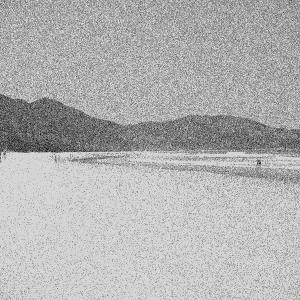} & \includegraphics[width=0.16\textwidth,trim=0in 0in 0in 0in,clip]{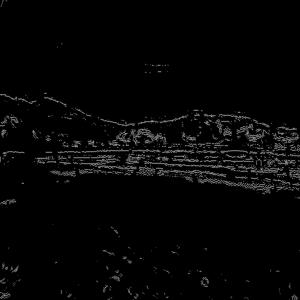} &  \includegraphics[width=0.16\textwidth,trim=0in 0in 0in 0in,clip]{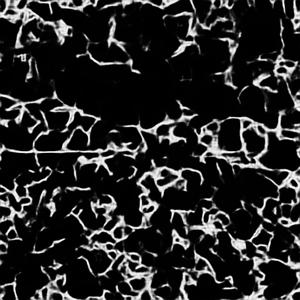} & \includegraphics[width=0.16\textwidth,trim=0in 0in 0in 0in,clip]{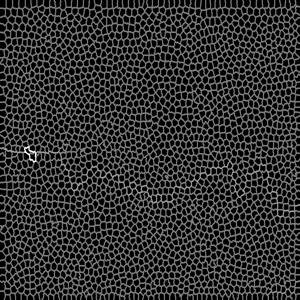} & \includegraphics[width=0.16\textwidth,trim=0in 0in 0in 0in,clip]{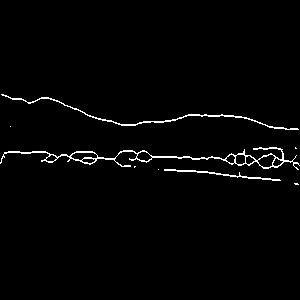} &\includegraphics[width=0.16\textwidth,trim=0in 0in 0in 0in,clip]{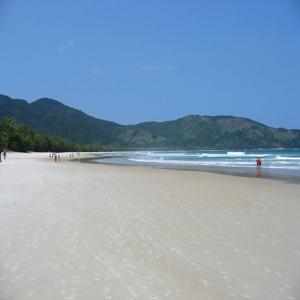}\\
    \includegraphics[width=0.16\textwidth,trim=0in 0in 0in 0in,clip]{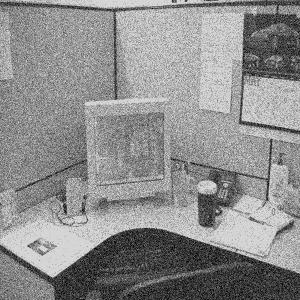} & \includegraphics[width=0.16\textwidth,trim=0in 0in 0in 0in,clip]{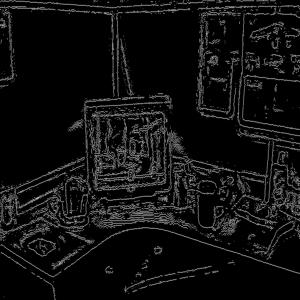} &  \includegraphics[width=0.16\textwidth,trim=0in 0in 0in 0in,clip]{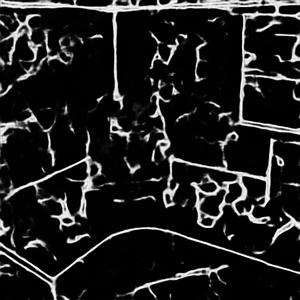} & \includegraphics[width=0.16\textwidth,trim=0in 0in 0in 0in,clip]{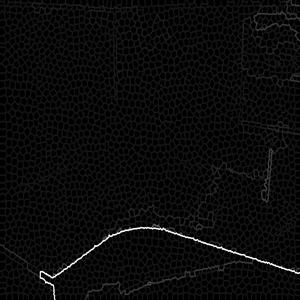} & \includegraphics[width=0.16\textwidth,trim=0in 0in 0in 0in,clip]{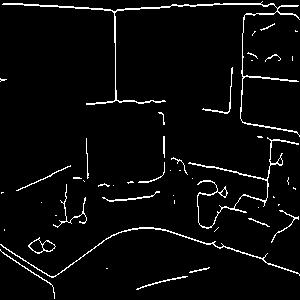} &\includegraphics[width=0.16\textwidth,trim=0in 0in 0in 0in,clip]{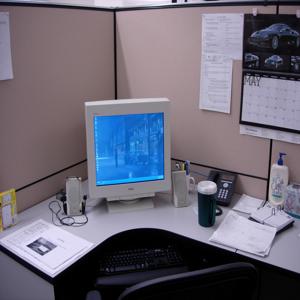}\\
    
     \scriptsize{Noisy Image} & \scriptsize{Low-res Edges} & \scriptsize{DexiNed} & \scriptsize{gPb}  & \scriptsize{Ours} & \scriptsize{Ground Truth}\\
\end{tabular}
\vspace{0.2cm}
\end{center}
	\caption{Examples of detecting the edges of noisy images. Gaussian noise was added to low-contrast images of letters (with $\sigma = 39\%$ of the range of the pixel values) and scenes (with $\sigma = 79\%$).
	Using SCFs we can make use of the low-resolution edges, by upscaling to the original size and completing the  broken contours. The resulting boundary recovers many more of the edges in the images. The same procedure of downscaling to remove noise did not improve the performance of the other edge-detection algorithms. See supplementary materials for more examples.} 
	\label{letters}
\end{figure*}

Another application of our SCF framework is to aid contour detection in images with high levels of noise. In cases where the image noise is extremely high, especially if edges are of low contrast, edge detection becomes extremely difficult. In Figure~\ref{letters}, we show images of a low contrast letter and a low contrast scene contaminated with noise.  In such cases, edge detection by classical algorithms \cite{canny1986computational,maire2008using}, and even state-of-the-art ones \cite{poma2020dense} fail.  Here, we propose a perceptual grouping based approach to detect such edges in the presence of noise. We first downscale the image. In cases where the noise is uncorrelated, reducing the image size will average out some of the noise, and allow some, but not all, of the edge to be detected. We detect edges using the logical-linear edge detector \cite{iverson1995logical}. 
The benefit of this edge detector is that it gives precise orientation information at each point. It can even give multiple orientations at a single point, allowing for the accurate representation of contour junctions and corners. The edge detector returns a map of edges and associated strengths in each direction. We binarize the map and trace the contours to obtain contour fragments that are each one pixel wide. After detecting edges we upscale the image. Up-scaling the image back to the original size will increase the number of missing contour segments. We make use of the orientation information given by the logical-linear edge detector as our keypoints. We then use SCFs to complete the gaps and yield a complete boundary.
To perform a more in-depth analysis of how well our edge detection performs in noise, we used the Artist dataset. In this dataset, color photographs from six categories of natural scenes (beaches, city streets, forests, highways, mountains, and offices) were downloaded from the internet and selected by Amazon MechanicalTurk workers looking for the most representative scenes for each category. Line drawings of these photographs were generated by trained artists at the Lotus Hill Research Institute  \cite{walther2011simple}. 
The resulting database had 475 line drawings in total
with 76-80 exemplars from each of the six categories: beaches, mountains, forests, highway scenes, city scenes, and office scenes. Images with fire or other potentially upsetting content were removed from the experiment (five images total). 
In Table \ref{fig:my_label}, we compare our method to the DexiNed \cite{poma2020dense} and gPb \cite{maire2008using} algorithms.

\section{Conclusion}

In this paper, we introduce a modernized SCF algorithm, and demonstrate its power when integrated with deep computer vision models. Our improvements include a method for automatically finding sources and sinks, so that our SCF algorithm no longer needs to have prior knowledge about the source and sink pairs for computation. In addition, we show that our algorithm mechanizes perceptual grouping processes, and mimics results in human perception when completing broken contour fragments. In addition to a Fokker-Planck equation implementation of the SCF algorithm, we introduce a 4-layer neural network model that computes SCF probability maps using a set of small kernel filters. We show how these implementations can be substituted into computer vision models that explicitly represent contour information. Specifically, we integrate our SCF blocks with inpainting generative models and models for edge-detection in noisy images. We find that our results outperform the state-of-the-art models with no additional training. Ultimately, our work improves upon the SCF algorithm and demonstrates how to integrate it in modern computer vision systems.
\newpage
\bibliography{egbib}
\end{document}